\documentclass{article}

\PassOptionsToPackage{numbers, compress}{natbib}
\usepackage[preprint]{neurips_2023}

\usepackage[utf8]{inputenc} % allow utf-8 input
\usepackage[T1]{fontenc}    % use 8-bit T1 fonts
\usepackage{hyperref}       % hyperlinks
\usepackage{url}            % simple URL typesetting
\usepackage{booktabs}       % professional-quality tables
\usepackage{amsfonts}       % blackboard math symbols
\usepackage{nicefrac}       % compact symbols for 1/2, etc.
\usepackage{microtype}      % microtypography
\usepackage[dvipsnames,table, xcdraw]{xcolor}         % colors
\usepackage{graphicx}
\usepackage{natbib}
\usepackage{pifont}% http://ctan.org/pkg/pifont
\usepackage{amssymb}% http://ctan.org/pkg/amssymb
\usepackage{amsmath}
\usepackage{cleveref}
\usepackage{enumitem}
\usepackage{longtable}

\title{Does Progress On Object Recognition Benchmarks Improve Real-World Generalization?}
\author{%
Megan Richards $^{1}$ \quad Polina Kirichenko $^{1,2}$ \quad Diane Bouchacourt $^1$ \quad Mark Ibrahim$^1$ \\
\\
$^1$ Meta AI (FAIR) \quad $^2$ New York University\\
}

\begin{document}

\maketitle

\begin{abstract}

For more than a decade, researchers have measured progress in object recognition on the ImageNet dataset along with its associated generalization benchmarks such as ImageNet-A, -C, and -R. Recent advances in foundation models, trained on orders of magnitude more data, have begun to saturate these standard benchmarks. Despite this progress, even today’s best models are brittle in practice. This suggests standard benchmarks, which tend to focus on predefined or synthetic alterations of images, may not be sufficient for measuring real world generalization. Consequently, we propose studying generalization across geography as a more realistic measure of progress using two datasets of objects from households across the globe. 
We conduct an extensive empirical evaluation of progress across nearly 100 vision models that span 16 architectures, including the most recent foundation models. 
We examine both the rate of progress and disparities in performance not revealed by average accuracy.
We first identify a progress gap between standard benchmarks and real-world, geographical shifts: progress on ImageNet results in up to $2.5$x more progress on standard generalization benchmarks than real-world distribution shifts. Second, we study model generalization across geographies by measuring the disparities in performance across regions, a more fine-grained measure of real world generalization. 
We observe all models have large geographic disparities, even foundation CLIP models, with differences of $7\% - 20\%$ in accuracy between regions. Counter to modern intuition, we discover progress on standard benchmarks fails to improve geographic disparities and in many cases exacerbates them: \textit{geographic disparities between the least performant models and today's best models have more than tripled}. Our results suggest scaling alone is insufficient for consistent robustness to real-world distribution shifts. Finally, we highlight in early experiments how simple last layer retraining on more representative, curated data can complement scaling as a promising direction of future work, reducing geographic disparity on both benchmarks by over two-thirds. 

\end{abstract}

\section{Introduction}

ImageNet \citep{russakovsky2015imagenet}, the standard benchmark for object recognition, has set the bar for progress in computer vision. 
Since its release in 2010, ImageNet along with other generalization benchmarks such as ImageNet-A,-C and -R \citep{imageneta,imagenetc,imagenetr} 
has spurred numerous advances in deep learning.
% \Megan{Saturated}
Now, more than a decade later, advances in scaling and multi-modal modeling have saturated these standard benchmarks. Most prominently, large scale vision-language models such as CLIP have been shown to achieve high-accuracies on in-distribution and other generalization benchmarks \citep{clip, clip_data_robustness, accuracyontheline}. 
% Recent work has even shown that this new generation of models are able to achieve nearly human-level performance on a benchmark of artificially distorted images \citep{human_machine_gap}. 
% \Megan{But best models are brittle in the real world}

Despite high performance on these standard benchmarks, model generalization remains an open problem — both vision and text models, as well as state-of-the-art (SOTA) multimodal models that take advantage of both, have been found to lack generalization abilities outside of those measured by standard benchmarks. Recent work has shown how CLIP \citep{clip} remains very vulnerable to changes in pose, background, size, position, and lighting \citep{ibrahim2022robustness, clipbrittle3Dlighting, clipbrittlepose, imagenete}. 
Such brittleness is however not reflect in standard ImageNet generalization benchmarks (also referred to as Out-of-Distribution, or OOD, benchmarks), as standard benchmarks focus on predefined or synthetic alterations of images that can not adequately reflect the rich diversity necessary for real world generalization \citep{imagenetc, imageneta, biasedcars}. 
We summarize commonly used generalization benchmarks in Table \ref{OOD-Summary}.

\begin{figure}[t!]
\centering
\includegraphics[width=\textwidth]{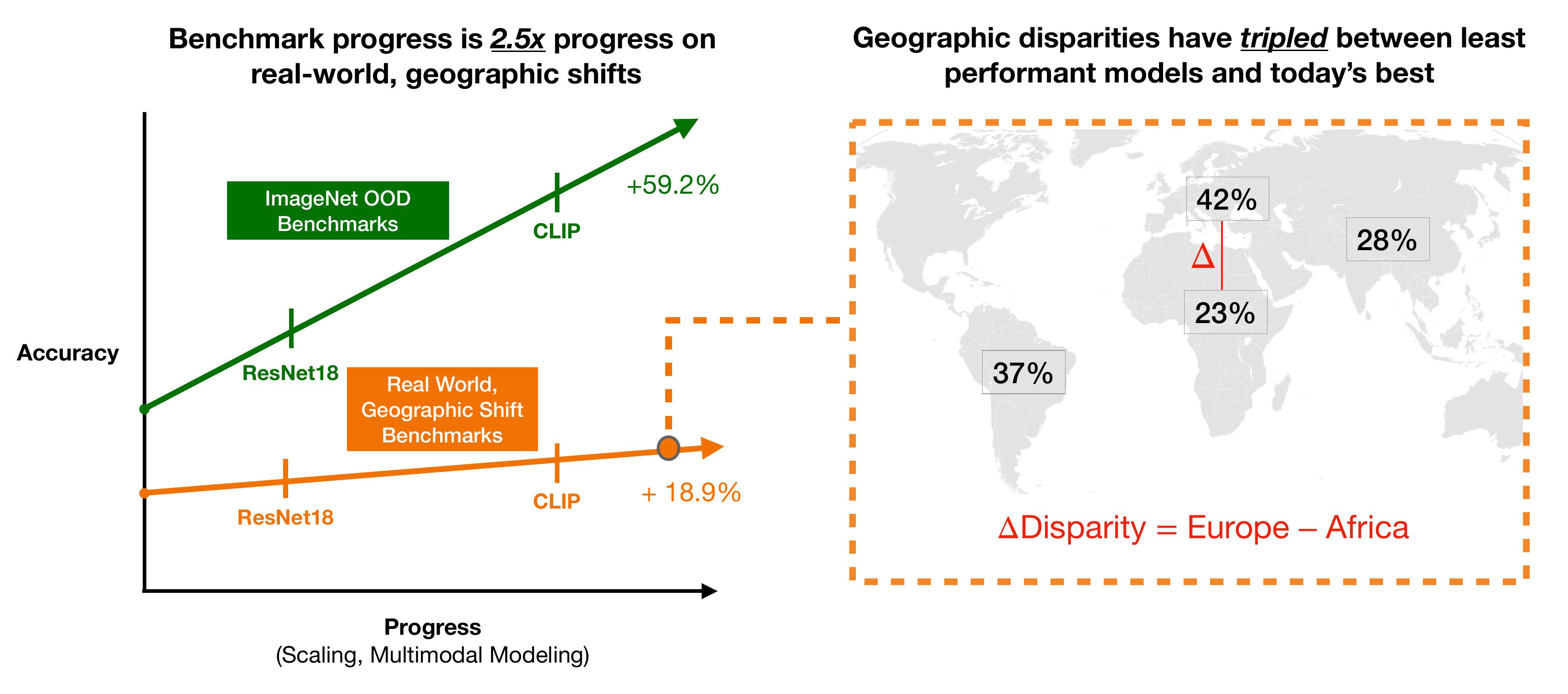}
\caption{\textbf{Progress rate on ImageNet generalization benchmarks is over 2.5x the progress rate on crowd-sourced, geographic shift benchmarks (\Cref{sec:inprogress})}. Further, geographic disparity between regions is exacerbated with progress on standard benchmarks, tripling over our range of models (\Cref{sec:gapprogress}).} 
\vspace{-5mm}
\label{fig:contribution_diagram}
\end{figure} 

As a step toward more realistic generalization measurement, we propose studying performance on crowd-sourced, globally representative datasets. We argue that such datasets offer two distinct advantages missing from current benchmarks. First, they allow us to assess models' performance under \textit{naturally occurring distribution shifts without simulated environments, preselected variations, or artificially injected transformations}. Second, they enable the measurement of geographic disparities between regions, a measure of generalization that is, by definition, relevant across the world, and a critical component of model safety in deployment that's often hidden in the commonly reported average accuracy. 
   
Using these datasets, we offer an extensive empirical analysis of generalization progress, evaluating nearly 100 vision models spanning 16 architectures, 8 distinct pretraining datasets, and a comprehensive set of foundation models. In addition, we systematically study of the impact of common robustness interventions as well as scaling of both model size and data. Our contributions are:
\begin{itemize}[noitemsep]
    \item In order to capture natural distribution shifts that affect real world applications, we propose to measure generalization via performance on globally crowd-sourced datasets (\Cref{sec:measure}).
    \item We identify a significant progress gap, finding progress on ImageNet results in up to $2.5$x progress on standard benchmarks than on real-world distribution shifts (Sec \Cref{sec:inprogress}). We illustrate this in the left part of Figure \ref{fig:contribution_diagram}. 
    \item We find, contrary to conventional wisdom, that improvements on standard benchmarks exacerbate generalization disparities across geographies: disparities in performance have \textit{tripled} between early models and today's best models (\Cref{sec:gapprogress}) as shown in the right part of Figure \ref{fig:contribution_diagram}.
    \item We study the impact of common robustness interventions and scaling, finding these two directions are not sufficient to close the geographic generalization gap. We explore curating more representative datasets as a promising path to mitigating the trade-offs we uncover (\Cref{sec:finetune}).
\end{itemize}

We hope our work will inspire researchers to look beyond standard benchmarks to improve real world generalization. To support future work, we will release our model test bed and evaluation code in a ready-to-use package, allowing researchers to run their own evaluations with just 4 lines of code.  

\begin{table}[t!]
\begin{tabular}{llccc}
\hline
Benchmark          & \multicolumn{1}{l}{shift type} & is natural                 & \# shift types & CLIP (ViT-L14)     \\ \hline
ImageNet           & \multicolumn{1}{c}{-}          & -                          & -                    & 76.2                 \\
ImageNet-V2        & \multicolumn{1}{c}{-}          & -                          & -                    & 70.1                 \\
ImageNet-Sketch    & drawing                        & $ \checkmark $             & 1                    & 60.2                 \\
ImageNet-Rendition & drawing                        & $ \checkmark $             & 1                    & 88.9                 \\
ObjectNet          & pose, background               & $ \checkmark $             & 3                    & 72.3                 \\
ImageNet-C         & corruptions                    & \ding{55} & 5              & 58.2      \\
ImageNet-A         & adversarial                    & $ \checkmark $             & 1                 & 77.1                \\ \hline
\textbf{DollarStreet}   & geographic                     & $ \checkmark $             & unlimited         & 17.0                 \\
\textbf{GeoDE}          & geographic                     & $ \checkmark $             & unlimited           & 6.5                 \\
                   &                                & \multicolumn{1}{l}{}       & \multicolumn{1}{l}{} & \multicolumn{1}{l}{}
\end{tabular}
    
  \caption{  
  \textbf{Geographic shift benchmarks enable measuring generalization to naturally occurring distribution shifts without simulated environments, preselected variations, or artificially injected transformations}. The last column represents the accuracy for each benchmark. For DollarSteet and GeoDE in the last column we show geographic disparity measured as the maximum performance difference between regions. CLIP numbers except ImageNet-C, DollarStreet, and GeoDE taken from \citep{clip}.}. 
  \label{OOD-Summary}
\end{table}

\section{Related work}

\paragraph{Generalization benchmarks do not fully reflect the real world}  

Real world generalization is a major challenge in deep learning. 
Consequently, a myriad of benchmarks were proposed to evaluate generalization capabilities of image classification models \citep{imagenetv2}. 
For example, ImageNet-A \citep{imageneta} was collected by intentionally mining challenging examples that fool a pre-selected model.
A complementary approach involves applying corruptions to images such as blurring, noise, or style alterations \citep{imagenetc,geirhos2018imagenet}.
Other benchmarks such as ImageNet-9 \citep{imagenet9}, ImageNet-R \citep{imagenetr}, \mbox{ImageNet-S} \citep{imagenetsketch} and ObjectNet \citep{objectnet} 
consist of images with a few predefined axes of generalization in mind: e.g.\ sketches in ImageNet-S or background variations in \mbox{ImageNet-9}. 
While being useful approximations of generalizations, such benchmarks either rely on artificially induced transformations or a predefined criteria 
that does not reflect the rich diversity of objects necessary for generalization in the real world \citep{degrave2021ai, robustness_synth_nat}.

\paragraph{Foundation models and robustness interventions} Many advances from robustness interventions to learning methods leveraging large scale data were proposed to improve generalization. 
Some robustness interventions are tailored to improve specific generalization axes such as to corruptions \citep{imagenetc}, texture \citep{geirhos2018imagenet}, or background shift\citep{ryali2021characterizing}. Data augmentation is a widely used technique which improves generalization \citep{pinto2023regmixup, augmix, cutmix, li2023whacamole}.
\citet{geirhos2020generalisation, robustness_spat_adv_tradeoff} and \citet{moayeri2022explicit} find that while robustness interventions improve generalization to the intended shift they may degrade performance to other shifts.
In parallel, self-supervised models \citep{goyal2022vision, shi2022robust} and more recent
foundation models \citep{pan2022contrastive} trained on much larger datasets (400M text-image pairs) show significant improvements on standard generalization benchmarks \citep{bommasani2022opportunities}.
However, in controlled synthetic settings, even large-scale foundation models were found to struggle with common variations in pose, background, and scale, among others \citep{abbas2022progress,ibrahim2022robustness,madan2023adversarial}. These results highlight that real-world out-of-distribution generalization still remains an open challenge.
% revealing more progress is necessary for the grand goal of real world generalization. 

\paragraph{The role of geography in classification} Geography presents a vital, real world axis for measuring generalization. 
Several classification datasets containing images from diverse geographic regions \citep{gustafson2023pinpointing, goyal2021self, goyal2022vision,rojasdollar,ramaswamy2023beyond} are used to study object classification models.
Their analysis reveals that classification models perform much better on some regions compared to others: accuracy gaps across regions can be as high as $20\%$ \citep{devries2019does}.
In conjunction, \citet{imagenet_western, imagenet_demographics, imagenet_content, shankar2017classification} present a possible explanation for this performance difference emphasizing over-representation of training images originating from Western geographies. 
Akin to \citet{DBLP:journals/corr/abs-2103-15796} which formulates geography as a benchmark for domain adaption, our work presents classification performance gaps across geographic regions 
as a benchmark of real world generalization progress.

\paragraph{Does better in-distribution performance lead to better out-of-distribution generalization?} 

\citet{chan2022data} shows that generalization in transformer models stems from aspects of the training distribution such as the number and rarity of training classes.
Specifically for foundation models such as CLIP, \citet{fang2022data,nguyen2023quality} show that the main factor driving improved generalization is the training data quality and distribution \citep{effective_robustness}.
\citet{accuracyontheline,agreementontheline} explicitly describe the relationship between ID and OOD showing ID performance is linearly correlated with OOD generalization.
Other work casts doubt on how well ID performance can predict real world OOD generalization \citep{imagenetv2, teney2022id}.
\citet{abnar2021exploring} show improved ID accuracy does not necessarily lead to downstream improvements.
\citet{imagenet_doesnt_transfer} show improvements on ID ImageNet classification does not lead to improvements on non-web scraped data.
Our work complements these studies by proposing classification gaps across geographies as an important, real-world marker of progress in generalization. 

\section{Measuring Real-World Generalization}
\label{sec:measure}

The ImageNet dataset has been an instrumental measure of progress for object recognition \citep{russakovsky2015imagenet}.
Alongside, standard ImageNet benchmarks such as ImageNet-A, ImageNet-C, and ObjectNet, have been developed to assess how well models generalize (see discussion in related work). 
With recent advances in foundation models such as CLIP however, performance (shown in Table \ref{OOD-Summary}) on the standard ImageNet distribution shifts benchmarks has begun to saturate, with best models achieving high accuracies matching that on original ImageNet. 
A limitation of standard ImageNet benchmarks is that they rely on artificially induced corruptions or predefined criteria
that can not adequately capture the rich diversity necessary for approximating generalization in the real world.

\subsection{Geographically diverse datasets}
\label{sub:diverse-datasets}

Recently, two datasets of household objects spanning the globe were introduced: DollarStreet \citep{rojasdollar} and GeoDE \citep{ramaswamy2023beyond}.
DollarStreet contains 38K images, with 96 classes, and spans 54 countries and 4 regions, while GeoDE contains 61K images with 40 classes, and spans 6 regions. 
Both datasets are commonly used in fairness literature to study performance disparities across images from different socioeconomic groups and regions \citep{devries2019does, gustafson2023pinpointing, rojasdollar, goyal2021self, goyal2022vision, ramaswamy2023beyond}. 
To study the largest catalogue of models possible, we use the ImageNet-1k mapping released for DollarStreet and generated a similar mapping for GeoDE. 
These class mappings (see \Cref{app:real-world}) allow us to 
evaluate any vision model compatible with the original 1k ImageNet classes. 
\textit{Geographically labeled datasets} such as GeoDE or DollarStreet allow us to measure generalization as it occurs in the real world across geographies.

\paragraph{Image quality and geographic representation} Can performance differences be simply be attributed to a lack of geographic representation or regional differences in image quality? 
As shown in \citet{ramaswamy2023beyond} and \citet{gustafson2023pinpointing} both DollarStreet and GeoDE have consistent image quality and contain roughly balanced numbers of samples per region.
In both datasets, images are crowd-sourced and labeled by the households who took the photo. This process produces high-quality ground truth class labels.

\subsection{Measuring generalization beyond average accuracy} 

The most commonly reported measure of progress for standard object recognition benchmarks is the \textit{average classification accuracy}.
We complement average accuracy with two additional metrics by assessing the rate of progress and uncovering disparities not revealed by average accuracy.

First, we are interested in measuring the rate at which each type of benchmarks (geographical or standard) benefit from advances in the field. 
  
Thus, we measure the rate of progress on each benchmark with respect to original ImageNet, where the rate of progress is the slope of a linear fit. We compute the difference of progress rates between standard generalization benchmarks and geographical shift benchmarks and consider ${\textbf{Progress Gap}}$ defined as:
\begin{flalign}
    {\textbf{Progress Gap}} &:= 
    {\mathrm{Progress\ Rate\ Standard - Progress\ Rate\ Geographical}}\\
    &=\frac{\mathrm{Standard\ Improvement - Geographic\ Improvement}}{\mathrm{ImageNet\ Improvement}}.
    \label{eq:progress-gap}
\end{flalign}

\textit{Progress gap} indicates how much of the progress on standard benchmarks transfers to real world geographic generalization. 
For example, a progress gap of 2x indicates improvements on standard benchmarks progress twice as fast as improvements on real world geographic generalization. 

However, there is a blind spot when using average accuracy: it may conceal poor performance on some groups relative to others \citep{pmlr-v177-idrissi22a}.
For example, a model may perform well on average, but generalize quite poorly to some regions.
Fortunately, datasets with geographic labels, such as DollarStreet and GeoDE, offer an opportunity to reveal when such disparities arise.

To complement average accuracy, we propose measuring \textit{geographic disparity} as an indicator of generalization in the real world. For DollarStreet and GeoDE, we do so by measuring the maximum absolute difference in a model's classification performance across any two regions, which we refer to as Geographic Disparity and is defined as:
\begin{equation}
    \Delta {\textbf{Disparity}} := \max \{| P_i - P_j| : i, j \in 1, \dots, k \}
    \label{eq:disparity}
\end{equation}
% \Diane{same here, don't like the italic of Geo Disparity}
where $P_i$ indicates the performance on the $i^{th}$ region and $k$ is the number of regions.
Of course, this definition can be applied broadly to any geographically labeled dataset and groupings 
other than regions such as country, zip code, or continent. 

Progress gap, together with geographical disparity in both GeoDE and DollarStreet datasets, gives us a more comprehensive assessment of real-world generalization in object recognition.

\subsection{Assessing progress in real-world generalization}

Equipped with two geographically diverse datasets and metrics of improvement, we now turn to the question: 
\textit{to what extent has progress on standard ImageNet benchmarks improved real world generalization?} First, we compare progress rates on standard benchmarks relative to progress based on average classification accuracy of household objects around the globe
(i.e. with ${\textbf{Progress Gap}}$ from Equation \ref{eq:progress-gap}). We go beyond average accuracy to probe how progress on standard benchmarks affects generalization in terms of geographic disparities (i.e. with $\Delta {\textbf{Disparity}}$ from Equation \ref{eq:disparity}) using DollarStreet and GeoDE described in \Cref{sub:diverse-datasets}.

We investigate a testbed of 98 models, which spans 16 architectures and includes recent foundation models such as CLIP, FLAVA, and DINOv2. We primarily use weights available in the Timm library \citep{rw2019timm} for ImageNet trained models, use the OpenCLIP library for CLIP models \citep{Ilharco_OpenCLIP_2021}, and use HuggingFace  \citep{huggingface} implementations of other foundation models. We include a comprehensive table of testbed metadata in \Cref{app:real-world}. 

\section{There is a Progress Gap Between Standard Benchmarks and Real-World Geographic Shifts}
\label{sec:inprogress}

Here we measure the rates of progress on standard and geographic benchmarks to study the extent to which progress on standard benchmarks improves real-world generalization.
If standard benchmarks faithfully reflect real world generalization, we would expect both benchmarks to have consistent rates of progress.
We compare the improvements on standard generalization benchmarks to geographic benchmarks as a function of ImageNet accuracy. As shown in \Cref{tab:progress-gap}, we find accuracy on standard generalization benchmarks to improve by 62.75\% on average, while progress on the geographically diverse DollarStreet dataset only improves by 18.9\% (33.5\% for GeoDE).

To isolate these progress trends, we compute linear trend lines for each benchmark. We find the trend lines are statistically significant with high Coefficients of Determination ($R^2$) as shown in \Cref{tab:progress-gap} (details in \Cref{app:progress-gap}). We discover a striking progress gap between standard generalization benchmarks and real world geographic shifts: \textit{progress on standard benchmarks is 2.5x the progress on real-world geographic shifts}. 
The progress gap is consistent for both DollarStreet and GeoDE, despite these benchmarks containing different classes and collection procedures. This suggests the progress gap isn't an artifact of a particular dataset. Both the difference in progress rates, and the net improvement values point to a substantial gap in progress between the commonly reported standard benchmarks and real-world geographic benchmarks.

\begin{table}[h!]
  \centering
  \begin{tabular}{lllll}
    \toprule
    Benchmark & Net Improvement($\uparrow$) & Progress Rate ($\uparrow$) & \textbf{${\textbf{Progress Gap}}$}  & $R^2$ ($\uparrow$)  \\
    \midrule
    DollarStreet (baseline) & \textcolor{OliveGreen}{+18.92\%} & 0.53 & 1.0x & 0.93 \\
    % GeoDE & \textcolor{OliveGreen}{+33.54\%}  & 1.03 & 0.95   \\
    
     \midrule 
     \textit{In-Distribution} & & & & \\
     ImageNet-V2 & \textcolor{OliveGreen}{+37.74\%}  &  1.18 & \textbf{2.2x}  & 0.99\\
    
    \midrule
    \textit{Out-Of-Distribution} & & & & 
    \vspace{0.1em}\\
    ImageNet-Sketch & \textcolor{OliveGreen}{+63.00\%}  & 1.37 & \textbf{2.6x}  & 0.75\\
    ImageNet-Rendition & \textcolor{OliveGreen}{+73.42\%} & 1.50 &\textbf{2.8x} & 0.74 \\
    ObjectNet & \textcolor{OliveGreen}{+51.84\%}  & 1.46 & \textbf{2.8x}  &  0.90 \\
   
    OOD Average & \textcolor{OliveGreen}{+62.75\%}  & 1.44 & \textbf{2.7x}  & 0.82\\
    \bottomrule
  \end{tabular}
     \label{tab:progress-gap}
    \caption{\textbf{There is a striking progress gap between standard ImageNet benchmarks and geographic shift benchmarks}, with all benchmarks improving at \textit{over double} the rate of DollarStreet. This translates to a net improvement on average OOD datasets that is more than \textit{3x} the net improvement on DollarStreet. We measure progress rate as the slope of a linear fit between ImageNet accuracy and benchmark accuracy, and include the coefficient of determination ($R^2$) for each.}
\end{table}

\section{Progress on Standard Benchmarks Exacerbates Performance Disparities }
\label{sec:gapprogress}

We found progress on real-world generalization in terms of average accuracy lags considerably behind progress on standard benchmarks.
While useful, average accuracy can conceal large disparities in performance indicative of poor geographic generalization.
Here we address average accuracy's blind spots by studying performance disparities across regions.
We measure performance disparity as the top-1 accuracy difference between the best (European) and least (Africa) performing regions in DollarStreet and GeoDE. 
We then study whether progress on standard ImageNet benchmarks improves or exacerbates geographic disparities.

\subsection{Even Today's Best Models Have Large Performance Disparities Between Regions}

We first measure the maximum performance disparity across regions. If a model generalizes well across geographies, we would expect a small performance disparity; whereas, poor geographic generalization would lead to large 
disparities.
We find all models have substantial disparities between regions, from ResNets to the largest CLIP models trained on 2 billion image-text pairs. 
In our study, ResNet models have average geographic disparities of 14.5\% on DollarStreet and 5.0\% on GeoDE. 
The best performing CLIP model actually had even more considerable disparities, with a disparity of 17.0\% on DollarStreet and 6.5\% on GeoDE. 
These considerable geographic disparities suggest average accuracy is concealing a crucial axes of generalization that remains \textit{unsolved by today's best models}. 
Next, we study how progress on standard ImageNet benchmarks has affected geographic disparities.

\subsection{Progress on ImageNet fails to resolve these disparities, often exacerbating them. }

\begin{figure}[htb]
\centering
\includegraphics[width=0.9\textwidth]{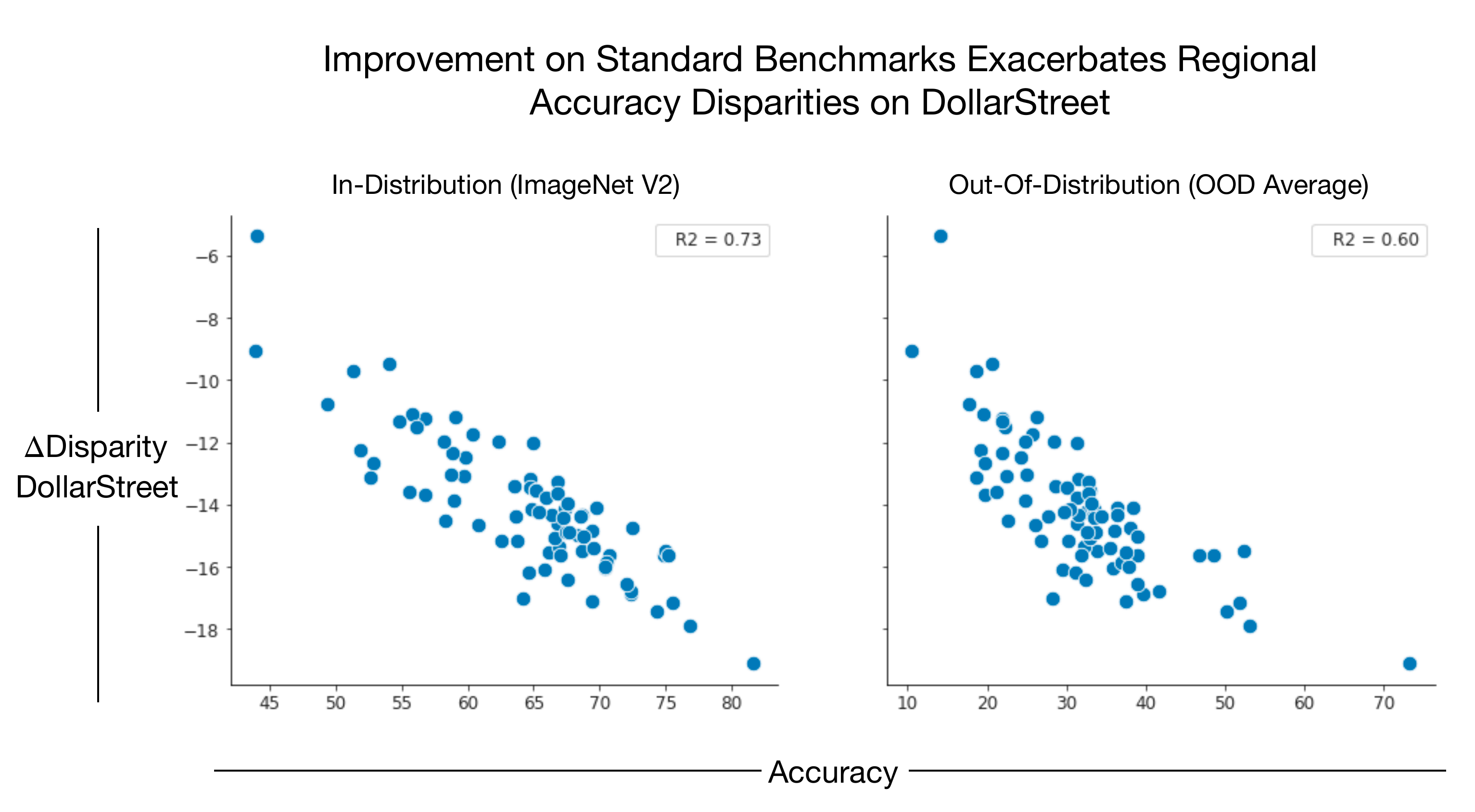}
\caption{\textbf{Model improvement on standard ID and OOD benchmarks exacerbates the region disparity on DollarStreet}, measured as the accuracy difference between Europe and Africa subsets.}
\label{fig:dollarstreet_grid}
\end{figure}

Has progress on standard ImageNet benchmarks improved or exacerbated geographic disparities?
To answer this question, we compare geographic disparities as a function of progress on ImageNet and standard generalization benchmarks.
Contrary to modern intuition, we discover, as shown in \Cref{fig:dollarstreet_grid}, progress on ImageNet and its generalization benchmarks not only fails to resolve geographic disparities, but actually exacerbates disparities.
We find for DollarStreet \textit{disparities between the least performant models and today’s best
models have more than tripled}. We also analyze performance disparities in GeoDE finding that 
that improvements on standard benchmarks are not predictive of any improvement in geographic disparity (see \Cref{app:performance-disparities}).

\subsection{Explaining the Widening Performance Disparities Between Regions}

\begin{figure}[htb]
\centering
\includegraphics[width=0.6\textwidth]{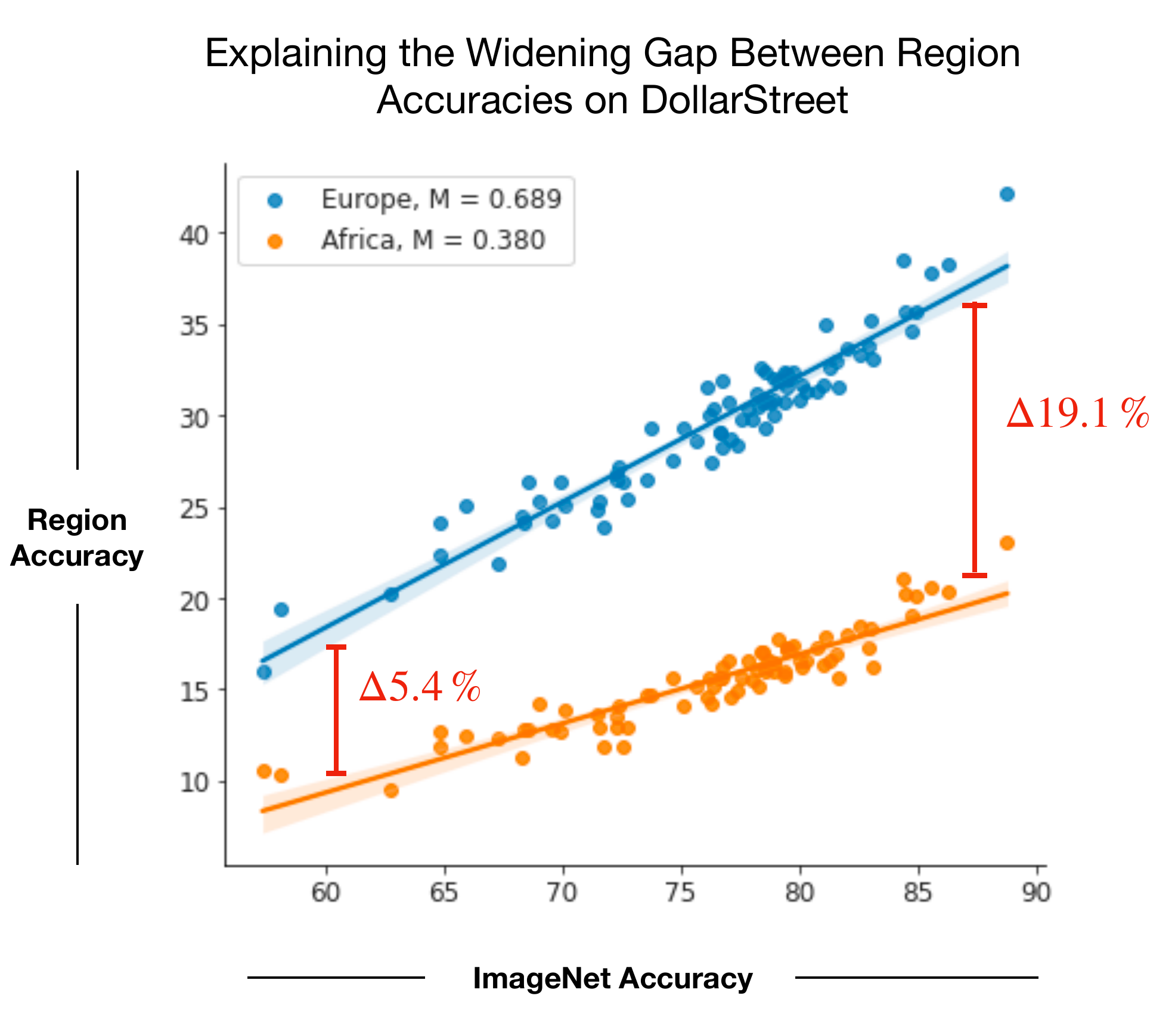}
\caption{\textbf{Model improvement on ImageNet exacerbates the region disparity on DollarStreet}, measured as the accuracy difference between Europe and Africa subsets.} 
\label{fig:explain_gap}
\end{figure} 

To explain the growing disparities, we isolate region performance as a function of improving ImageNet accuracy to understand individual effect on the rate of progress in each region.
In \Cref{fig:explain_gap}, we show accuracy in the best (Europe) and least (Africa) performing regions as ImageNet accuracy improves. 
While overall models also improve on each region, they improve on for Europe at almost double the rate of that for Africa, leading to a widening performance disparity between them. 
For GeoDE, we see much more similar rates of improvement across regions (see \Cref{app:performance-disparities}).

Our analysis indicates that progress as measured by average accuracy is an incomplete picture. We find that models across architectures and datasets have large, meaningful disparities between regions, and that improvement on current benchmarks fails to improve on these disparities. 

\section{Generalization across geography: open challenges and promising directions}
\label{sec:finetune}

Next, we explore promising directions for improving real-world generalization across geographies.
We investigate multiple avenues from common robustness interventions such MixUp to scaling of both data/model size as well as forms of data curation.
We find many avenues known to improve generalization on standard benchmarks fail to address real world geographic shifts.
Finally, we discover promise via simple last layer retraining \cite{kirichenko2022layer} on curated representative data for improving real-world geographic generalization. 

\subsection{Robustness interventions offer limited improvements}

We evaluate popular interventions that have been shown to improve generalization on standard benchmarks: Deep AugMix, AugMix, Texture Debiasing, CutMix, and AntiAliasing. We evaluate these techniques using pretrained ResNet50 models. In Table \ref{tab:robustness_interventions}, we show accuracy on standard benchmarks as well as geographic disparities for DollarStreet and GeoDE for models trained with each intervention compared to  
a baseline ResNet50 model (trained without any interventions). 
The majority of robustness interventions improved one benchmark's regional gap slightly, while degrading the other. 
The exception is AugMix, which improved the GeoDE and DollarStreet gaps by 1.86\% and 0.94\% respectively. 
Common robustness interventions overall offer limited improvements to real-world geographic disparities, indicating a need 
for more targeted solutions.

\begin{table}[h!]
\setlength\tabcolsep{2pt}
  \centering
  \begin{tabular}{lllll}
    \toprule
    Intervention  & ImageNet $(\uparrow)$ & OOD Avg $(\uparrow)$ & $\Delta \textbf{Disparity}$  GeoDE  $(\downarrow)$ & $\Delta \textbf{Disparity}$ DS $(\downarrow)$  \\
    \midrule
  \textcolor{gray}{Baseline}  & \textcolor{gray}{76.34} & \textcolor{gray}{30.28} & \textcolor{gray}{4.96} & \textcolor{gray}{15.16} \\
   Deep AugMix  & 76.73 & 32.92 & 5.22 & 13.53 \\
    Texture Debiased  & 76.73 & 31.13 & 4.70 & 16.20 \\
    Ant-Aliased & 77.41 & 30.09 & 5.54 & 13.46 \\
    AugMix & 77.53 & 32.51 & 3.10 & 14.22 \\
    CutMix & 78.58 & 29.43 & 4.38 & 16.10 \\
    \bottomrule
  \end{tabular}
    \caption{Benchmarking Robustness Interventions. Most robustness interventions produced mixed results, with the exception of AugMix, which provided small improvements to geographic disparities and ImageNet accuracy. DS refers to DollarStreet.}
  \label{tab:robustness_interventions}
\end{table}

\subsection{Foundation Vision Models and Scaling}

\begin{figure}[htb]
\centering
\includegraphics[width=0.9\textwidth]{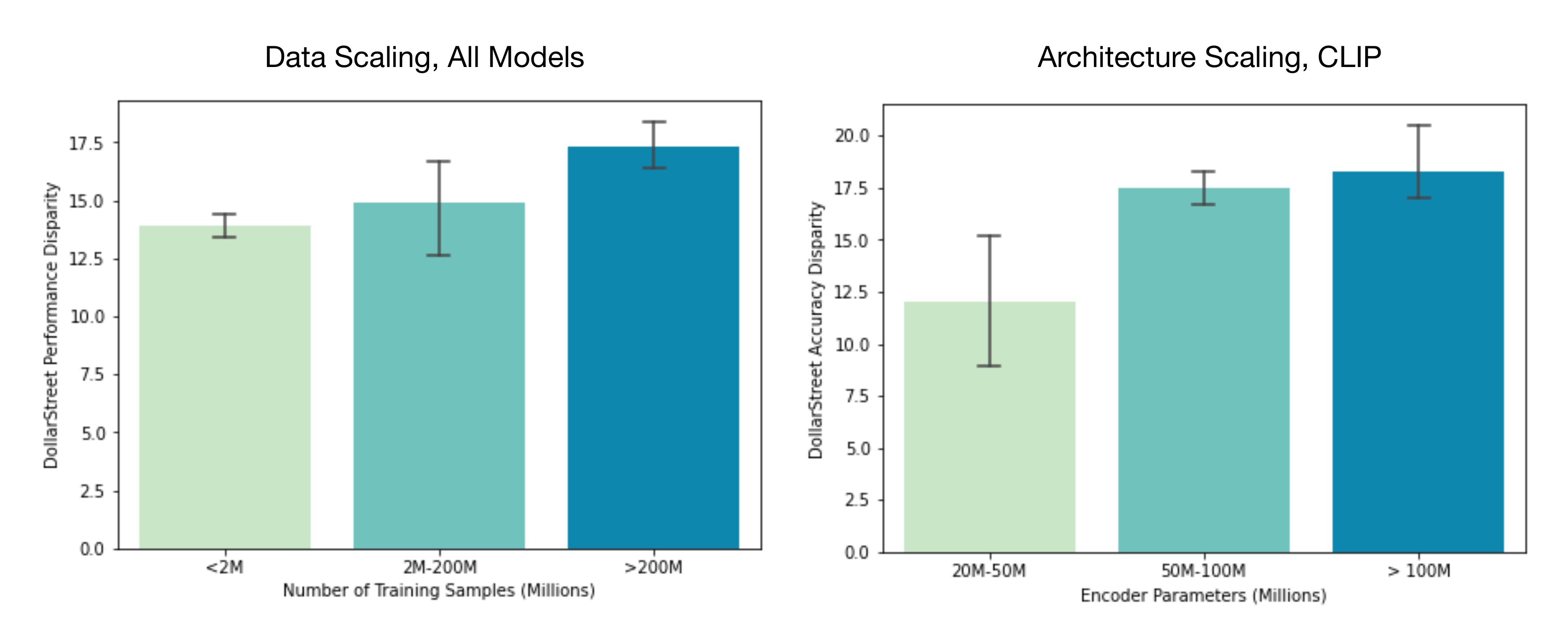}
\caption{\textbf{Dataset and architecture scaling exacerbates region disparities on DollarStreet.}}
\label{fig:scale_clip}
\end{figure}

Scaling of both model size and training data have been successful strategies behind many recent advances \citep{clip}.
Here we study whether scaling's success on standard benchmarks translates to progress on real-world geographic generalization.
We measure geographic disparity $\Delta \textbf{Disparity}$ as a function of scale in terms of data (+200 million) and model size (+100 million parameters) in Figure \ref{fig:scale_clip}.
We find \textit{neither scaling data nor model size improves geographic disparities}.
While error bars don't allow us to draw any conclusive trends, in terms of averages scaling both model and data sizes seems to exacerbate geographic disparities.
We replicate the GeoDE plots in Appendix \ref{app:foundation_and_scaling}, which contain the same relationship. We also show the scaling trend per architecture type in Appendix \ref{app:foundation_and_scaling}, but did not find any promising scaling trends by architecture. 

Our results suggest scaling alone is insufficient for robustness to real-world distribution shifts. Even CLIP models have these persistent performance disparities between regions, which are not mitigated by scaling CLIP. 
\subsection{The Promise of Curating Representative Balanced Data}

Finally, we explore careful data curation as a promising direction. 
Prior work has highlighted data quality as a critical component of robustness improvements \cite{clip_data_robustness, pmlr-v177-idrissi22a}. Recent work has also found that careful data pruning can help surpass existing performance scaling laws \cite{beyondneuralscalinglaws}. In turn, we ask: to what extent can curating balanced, representative training data address geographic distribution shifts? We take a first step to answering this question by 1) analyzing the performance of DINOv2, a recent self-supervised foundation model trained with auto-curated video data, and 2) last layer retraining \citep{kirichenko2022layer} of ImageNet-pretrained ViT model on DollarStreet data.

\textbf{DINOv2} \quad Despite being a mid-size model at 86 million parameters, DINOv2 achieved the smallest GeoDE region performance disparity of our testbed, with just a 2.46\% accuracy difference between Europe and Africa subsets. While the model still had a significant region disparity on DollarStreet, the GeoDE improvement is remarkable for its size, and highlights that data curation offers a promising path to mitigating the tradeoff between geographic performance disparity and standard benchmarks.  

\paragraph{Last-layer retraining on geographically representative data} Do we need to retrain a model from scratch to reap the benefits of curated data? 
To answer this question, we implement the last-layer retraining method from \citet{kirichenko2022layer}, which retrains only the last layer of a pretrained model on the relevant downstream task. 
Here we retrain the last linear head of a ViT model \citep{dosovitskiy2020image} on the training split of DollarStreet.  
We train the last layer for 5 epochs using Adam optimizer, learning rate $10^{-5}$ and batch size 32. 
We then evaluate this model on both DollarStreet and GeoDE. For GeoDE, we evaluate on the subset of classes overlapping with the 1-k ImageNet classes
(full details in Appendix \ref{app:representative_data}. 
We find, as shown in \Cref{tab:finetuning}, last-layer retraining improves average accuracy and geographic disparities on both DollarStreet and GeoDE.
The average accuracy on DollarStreet's evaluation set improves by a dramatic 53.4\% with geographic disparity also improving by 11.7\%. 
Remarkably, despite GeoDE containing a different set of classes and retraining only on DollarStreet, we observe improvements on GeoDE of 11.5\% on average accuracy and 3.2\% in geographic disparities.
Our results indicate careful use of more representative data holds great promise to consistently improve both the average performance and geographic disparity of object recognition models. 

\begin{table}[h!]
  \label{tab:finetuning}
  \centering
    \setlength\tabcolsep{5pt}
  \begin{tabular}{lllll}
    \toprule 
    & \multicolumn{2}{c}{\textbf{Average Accuracy ($\uparrow)$}} & \multicolumn{2}{c}{$\Delta$ \textbf{Disparity} ($\downarrow)$}  \vspace{0.4em}\\
     & DollarStreet & GeoDE  & DollarStreet & GeoDE  \\
    \midrule
    \textcolor{gray}{ViT} & \textcolor{gray}{23.46} &\textcolor{gray}{65.44} & \textcolor{gray}{17.12} & \textcolor{gray}{4.86} \\
   LLR-ViT & 76.84 \tiny{$\pm$ 0.1} \normalsize{\textbf{\textcolor{OliveGreen}{(+53.41)}}} & 76.97 \tiny{$\pm$ 0.9} \normalsize{\textbf{\textcolor{OliveGreen}{(+11.53)}}} &  5.47 \tiny{$\pm$ 1.2} \normalsize{\textbf{\textcolor{OliveGreen}{(-11.65)}}} & 1.64 \tiny{$\pm$ 0.6} \normalsize{\textbf{\textcolor{OliveGreen}{(-3.22)}}}\\
    \bottomrule
  \end{tabular}
  \vspace{1em}
    \caption{\textbf{Last layer retraining on DollarStreet improves geographic disparity and overall performance on \textbf{both} DollarStreet and GeoDE.} As explained in text, we report GeoDE overlapping with ImageNet. LLR-ViT refers to Last-Layer Retrained ViT.}
\end{table}

\section{Discussion}
\label{sec:discussion}

In this paper, we explored generalization across geography as a more realistic measure of progress in object recognition. Not only does geography capture generalization across real world distribution shifts, it also realigns generalization with challenges seen in practice. Using two large, crowd-sourced global datasets, we identified a substantial progress gap: advances on ImageNet improve progress on standard benchmarks 2.5x the progress on real-world geographic generalization. Moreover, while today's best models substantially improve standard benchmarks, they actually exacerbate geographic disparities
by nearly $3$ times. For this reason, we highlighted the importance of including
more benchmarks that reflect real-world challenges, on which progress might not align with standard ImageNet benchmarks.
We showcased the promise of using more curated and/or representative data for solving geographic disparities.
We will release our code and test bed to encourage the research community to make progress on real-world generalization. In future work, we would like to extend our study to other axes of real-world shifts.

\bibliography{biblio}
\bibliographystyle{plainnat}

\newpage
\appendix

\section{Appendix}

\subsection{Measuring Real-World Generalization}
\label{app:real-world}

\paragraph{TestBed and Evaluation Procedure:} We include a list of the models in our testbed below, including the architecture group, evaluation type, training dataset, and the library or github source we used for model weights. For data augmentation, for all models we used the ImageNet normalization available in PyTorch, resize images to 256 pixels, and center crop to 224 pixels. 

\begin{longtable}{lllll}
\hline
\textbf{Model} & \textbf{Architecture} & \textbf{Evaluation Type} & \textbf{Dataset} & \textbf{Source} \\ \hline
\endfirsthead
\endhead
eva-clip                   & CLIP                  & 1K                    & Laion-2B         & Timm            \\
convnext-base              & ConvNext              & 1K                    & 1K               & Timm            \\
convnext-large             & ConvNext              & 1K                    & 1K               & Timm            \\
convnext-small             & ConvNext              & 1K                    & 1K               & Timm            \\
dla102                     & DLA                   & 1K                    & 1K               & Timm            \\
dla102x                    & DLA                   & 1K                    & 1K               & Timm            \\
dla169                     & DLA                   & 1K                    & 1K               & Timm            \\
dla34                      & DLA                   & 1K                    & 1K               & Timm            \\
dla46c                     & DLA                   & 1K                    & 1K               & Timm            \\
dla46xc                    & DLA                   & 1K                    & 1K               & Timm            \\
dla60                      & DLA                   & 1K                    & 1K               & Timm            \\
dla60x                     & DLA                   & 1K                    & 1K               & Timm            \\
edgenet-base               & EdgeNext              & 1K                    & 1K               & Timm            \\
edgenet-s                  & EdgeNext              & 1K                    & 1K               & Timm            \\
edgenet-xs                 & EdgeNext              & 1K                    & 1K               & Timm            \\
edgenet-xxs                & EdgeNext              & 1K                    & 1K               & Timm            \\
hrnet18                    & HRNet                 & 1K                    & 1K               & Timm            \\
hrnet18small               & HRNet                 & 1K                    & 1K               & Timm            \\
hrnet30                    & HRNet                 & 1K                    & 1K               & Timm            \\
hrnet32                    & HRNet                 & 1K                    & 1K               & Timm            \\
hrnet40                    & HRNet                 & 1K                    & 1K               & Timm            \\
hrnet44                    & HRNet                 & 1K                    & 1K               & Timm            \\
hrnet48                    & HRNet                 & 1K                    & 1K               & Timm            \\
hrnet64                    & HRNet                 & 1K                    & 1K               & Timm            \\
lcnet100                   & LCNet                 & 1K                    & 1K               & Timm            \\
lcnet50                    & LCNet                 & 1K                    & 1K               & Timm            \\
lcnet75                    & LCNet                 & 1K                    & 1K               & Timm            \\
mlpmixer                   & MLP                   & 1K                    & 1K               & Timm            \\
mlpmixerlarge              & MLP                   & 1K                    & 1K               & Timm            \\
mobilenet-lamb100          & MobileNet-V3          & 1K                    & 1K               & Timm            \\
mobilenet-lamb50           & MobileNet-V3          & 1K                    & 1K               & Timm            \\
mobilenet-lamb75           & MobileNet-V3          & 1K                    & 1K               & Timm            \\
regnet                     & RegNet                & 1K                    & 1K               & Timm            \\
regnet120                  & RegNet                & 1K                    & 1K               & Timm            \\
regnet16                   & RegNet                & 1K                    & 1K               & Timm            \\
regnet2                    & RegNet                & 1K                    & 1K               & Timm            \\
regnet32                   & RegNet                & 1K                    & 1K               & Timm            \\
regnet320                  & RegNet                & 1K                    & 1K               & Timm            \\
regnet6                    & RegNet                & 1K                    & 1K               & Timm            \\
regnet64                   & RegNet                & 1K                    & 1K               & Timm            \\
regnet8                    & RegNet                & 1K                    & 1K               & Timm            \\
seer1280                   & RegNet                & 1K                    & Instagram        & Github$^*$            \\
seer320                    & RegNet                & 1K                    & Instagram        & Github$^*$             \\
seer640                    & RegNet                & 1K                    & Instagram        & Github$^*$              \\
regnet120x                 & RegNetX               & 1K                    & 1K               & Timm            \\
regnet16x                  & RegNetX               & 1K                    & 1K               & Timm            \\
regnet2x                   & RegNetX               & 1K                    & 1K               & Timm            \\
regnet320x                 & RegNetX               & 1K                    & 1K               & Timm            \\
regnet32x                  & RegNetX               & 1K                    & 1K               & Timm            \\
regnet4x                   & RegNetX               & 1K                    & 1K               & Timm            \\
regnet64x                  & RegNetX               & 1K                    & 1K               & Timm            \\
regnet6x                   & RegNetX               & 1K                    & 1K               & Timm            \\
regnet8x                   & RegNetX               & 1K                    & 1K               & Timm            \\
resnet101                  & ResNet                & 1K                    & 1K               & Timm            \\
resnet152                  & ResNet                & 1K                    & 1K               & Timm            \\
resnet18                   & ResNet                & 1K                    & 1K               & Timm            \\
resnet34                   & ResNet                & 1K                    & 1K               & Timm            \\
resnet50                   & ResNet                & 1K                    & 1K               & Timm            \\
resnet50anti               & ResNet                & 1K                    & 1K               & Timm            \\
resnet50augmix             & ResNet                & 1K                    & 1K               & Timm            \\
resnet50cutmix             & ResNet                & 1K                    & 1K               & Timm            \\
resnet50cutmixbaseline     & ResNet                & 1K                    & 1K               & Timm            \\
resnet50deepaug            & ResNet                & 1K                    & 1K               & Timm            \\
resnet50deepaugmix         & ResNet                & 1K                    & 1K               & Timm            \\
resnet50texture            & ResNet                & 1K                    & 1K               & Timm            \\
rexnet100                  & RexNet                & 1K                    & 1K               & Timm            \\
rexnet130                  & RexNet                & 1K                    & 1K               & Timm            \\
rexnet150                  & RexNet                & 1K                    & 1K               & Timm            \\
rexnet200                  & RexNet                & 1K                    & 1K               & Timm            \\
tinynet-a                  & TinyNet               & 1K                    & 1K               & Timm            \\
tinynet-b                  & TinyNet               & 1K                    & 1K               & Timm            \\
tinynet-c                  & TinyNet               & 1K                    & 1K               & Timm            \\
tinynet-e                  & TinyNet               & 1K                    & 1K               & Timm            \\
vgg-11                     & VGG                   & 1K                    & 1K               & Timm            \\
vgg-13                     & VGG                   & 1K                    & 1K               & Timm            \\
vgg-16                     & VGG                   & 1K                    & 1K               & Timm            \\
vgg-19                     & VGG                   & 1K                    & 1K               & Timm            \\
DINOv2                 & ViT                   & 1K                    &  LVD-142M          & GitHub$^{**}$      \\
vit                        & ViT                   & 1K                    & 21K              & Timm            \\
vitlarge                   & ViT                   & 1K                    & 21K              & Timm            \\
clip-convnext-laion2b      & CLIP                  & Zeroshot              & Laion-2B         & OpenCLIP        \\
clip-convnext-laion2b-a    & CLIP                  & Zeroshot              & Laion-2B         & OpenCLIP        \\
clip-convnext-laion2b-aug  & CLIP                  & Zeroshot              & Laion-2B         & OpenCLIP        \\
clip-convnextlarge-laion2b & CLIP                  & Zeroshot              & Laion-2B         & OpenCLIP        \\
clip-r101-openai           & CLIP                  & Zeroshot              & OpenAI           & OpenCLIP        \\
clip-r101-yfcc             & CLIP                  & Zeroshot              & YFCC                 & OpenCLIP        \\
clip-r50-cc12m             & CLIP                  & Zeroshot              &  CC12M                & OpenCLIP        \\
clip-r50-openai            & CLIP                  & Zeroshot              & OpenAI           & OpenCLIP        \\
clip-r50-yfcc              & CLIP                  & Zeroshot              &  YFCC                & OpenCLIP        \\
clip-vit14-laion2b         & CLIP                  & Zeroshot              & Laion-2B         & OpenCLIP        \\
clip-vit14-laion400m       & CLIP                  & Zeroshot              & Laion-400M       & OpenCLIP        \\
clip-vit14-openai          & CLIP                  & Zeroshot              & OpenAI           & OpenCLIP        \\
clip-vit16-laion2b         & CLIP                  & Zeroshot              & Laion-2B         & OpenCLIP        \\
clip-vit16-laion400m       & CLIP                  & Zeroshot              & Laion-400M       & OpenCLIP        \\
clip-vit16-openai          & CLIP                  & Zeroshot              & OpenAI           & OpenCLIP        \\
clip-vit32-laion400m       & CLIP                  & Zeroshot              & Laion-400M       & OpenCLIP        \\
clip-vit32-openai          & CLIP                  & Zeroshot              & OpenAI           & OpenCLIP        \\
flava                      & PMD                  & Zeroshot              & HuggingFace            & OpenCLIP       
\end{longtable}
$^*$ The SEER Github can be found here: \url{https://github.com/facebookresearch/vissl/tree/main/projects/SEER}.   \\
$^{**}$The DINOv2 Github can be found here: \url{https://github.com/facebookresearch/dinov2}. 

\paragraph{Class Maps}
For DollarStreet and GeoDE datasets, we use a class mapping to ImageNet-1K to evalute 1K models, and use the original labels for DollarStreet and GeoDE to evalaute zero-shot models. We use the released mapping for DollarStreet and generate mapping for GeoDE. We generate the GeoDE mapping using the spacey model \citep{spacey_package} to calculate the most similar ImageNet classes for each GeoDE class, manually selecting the most reasonable results and correcting as needed. We successfully create mappings for 36 of the 40 GeoDE classes.  Below are the class mappings:

\begin{longtable}{ll}
\label{ds_class_map}
\textbf{DollarStreet Class}          & \textbf{ImageNet Class(es)} \\ \hline
\endfirsthead
\endhead
home                                 & manufactured home           \\
street view                          & street sign                 \\
tv                                   & television                  \\
washing clothes/cleaning             & washing machine             \\
toilet                               & toilet seat                 \\
kitchen sink                         & washbasin                   \\
drinking water                       & water bottle                \\
stove/hob                            & stove                       \\
salt                                 & salt shaker                 \\
bed                                  & day bed                     \\
toys                                 & toyshop                     \\
everyday shoes                       & running shoe                \\
plate of food                        & plate                       \\
cooking pots                         & skillet                     \\
social drink                         & soda bottle                 \\
phone                                & cellphone                   \\
place where eating dinner            & dining table                \\
lock on front door                   & padlock                     \\
wardrobe                             & wardrobe                    \\
soap for hands and body              & soap dispenser              \\
ceiling                              & tile roof                   \\
refrigerator                         & refrigerator                \\
bathroom/toilet                      & toilet seat                 \\
dish washing brush/cloth             & dishrag                     \\
toilet paper                         & toilet paper                \\
plates                               & plate                       \\
dish washing soap                    & soap dispenser              \\
trash/waste                          & trash can                   \\
dish racks                           & plate rack                  \\
shower                               & shower curtain              \\
cups/mugs/glasses                    & cup                         \\
armchair                             & rocking chair               \\
light sources                        & table lamp                  \\
light source in livingroom           & table lamp                  \\
books                                & bookcase                    \\
switch on/off                        & switch                      \\
light source in kitchen              & table lamp                  \\
couch                                & studio couch                \\
sofa                                 & studio couch                \\
roof                                 & tile roof                   \\
cutlery                              & wooden spoon                \\
cooking utensils                     & spatula                     \\
medication                           & medicine cabinet            \\
source of cool                       & electric fan                \\
pen/pencils                          & ballpoint                   \\
street detail                        & street sign                 \\
turning lights on and off            & switch                      \\
music equipment                      & speaker                     \\
tools                                & tool kit                    \\
cleaning equipment                   & dishrag                     \\
bed kids                             & day bed                     \\
table with food                      & dining table                \\
get water                            & water jug                   \\
paper                                & paper towel                 \\
radio                                & radio                       \\
shoes                                & running shoe                \\
starting stove                       & igniter                     \\
freezer                              & icebox                      \\
source of heat                       & space heater                \\
computer                             & desktop computer            \\
jewelry                              & necklace                    \\
knifes                               & paper knife                 \\
wall clock                           & wall clock                  \\
pouring water                        & water jug                   \\
doing dishes                         & dishwasher                  \\
guest bed                            & day bed                     \\
mosquito protection                  & mosquito net                \\
bike                                 & all-terrain bike            \\
pouring drinking water               & water bottle                \\
oven                                 & stove                       \\
place where serving guests           & eating place                \\
glasses or lenses                    & dark glasses                \\
necklaces                            & necklace                    \\
source of light                      & table lamp                  \\
parking lot                          & parking meter               \\
waste dumps                          & trash can                   \\
eating                               & restaurant                  \\
car                                  & passenger car               \\
reading light                        & table lamp                  \\
lightsources by bed                  & table lamp                  \\
family eating                        & eating place                \\
arm watch                            & digital watch               \\
taking a teaspoon of salt            & salt shaker                 \\
using toilet                         & toilet seat                 \\
sitting and watching tv              & television                  \\
opening and closing the freezer      & icebox                      \\
diapers (or baby-pants)              & diaper                      \\
moped/motorcycle                     & moped                       \\
cleaning after toilet                & toilet paper                \\
dishwasher                           & dishwasher                  \\
opening and closing the refrigerator & refrigerator                \\
answering the phone                  & mobile phone                \\
alarm clock                          & analog clock                \\
wheel barrow                         & wheelbarrow                 \\
listening to the radio               & radio                       \\
dinner guests                        & eating place               
\end{longtable}

\begin{center}
\begin{longtable}{ll}
\hline
\textbf{GeoDE Class} & \textbf{ImageNet Class(es)} \\ \hline
\endfirsthead
\endhead
bag & backpack, purse, punching bag, sleeping bag, plastic bag, messenger bag, \\
& shopping basket, pencil case \\
hand soap & soap dispenser, lotion \\
dustbin & bucket, trash can, plastic bag, barrel \\
toothbrush & - \\
toothpaste toothpowder & - \\
hairbrush comb & - \\
chair & barber chair, folding chair, rocking chair, couch, throne \\
hat & cowboy hat, swimming cap, football helmet, poke bonnet, sombrero 
 \\ & military hat (bearskin or shako), shower cap \\
light fixture & table lamp, spotlight, lampshade, candle \\
light switch & electrical switch \\
plate of food & plate, tray \\
spices & - \\
stove & Dutch oven, stove \\
cooking pot & frying pan, hot pot, Crock Pot, cauldron, Dutch oven, wok \\
cleaning equipment & vacuum cleaner, washing machine, mop, broom, bucket, soap dispenser \\
lighter & lighter \\
medicine & pill bottle, medicine cabinet \\
candle & candle \\
toy & teddy bear, toy store \\
jug & water jug, whiskey jug, water bottle, drink pitcher \\
streetlight lantern & torch, pole \\
front door & sliding door \\
tree & - \\
house & cliff dwelling, mobile home, barn, home theater, boathouse \\
backyard & patio \\
truck & garbage truck, semi-trailer truck, tow truck, pickup truck \\
waste container & plastic bag, trash can, barrel, bucket \\
car & garbage truck, recreational vehicle, semi-trailer truck, tow truck, sports car, railroad car, \\
& minivan, station wagon, minibus, jeep, limousine, taxicab, convertible, pickup truck \\
&  moving van, police van, race car \\
fence & chain-link fence, picket fence, split-rail fence \\
road sign & traffic or street sign \\
dog & Bernese Mountain Dog, Sealyham Terrier, Toy Poodle, toy terrier, African wild dog, husky, \\ 
& Maltese, Beagle, Labrador Retriever, Cairn Terrier, dingo, Australian Kelpie \\ 
& German Shepherd Dog, Golden Retriever, Malinois, Norwegian Elkhound, Chihuahua, \\
& Tibetan Mastiff, Staffordshire Bull Terrier, American Staffordshire Terrier \\
&  Pembroke Welsh Corgi, Miniature Poodle, Basenji, Rhodesian Ridgeback,
\\ 
& Appenzeller Sennenhund, Ibizan Hound \\
wheelbarrow & wheelbarrow \\
religious building & mosque, church, monastery, bell tower, altar \\
stall & - \\
boat & motorboat, canoe, fireboat, lifeboat, sailboat, submarine, ocean \\ & liner, trimaran, catamaran \\
monument & triumphal arch, obelisk, stupa, pedestal, brass memorial plaque, megalith \\
flag & flagpole \\
bus & minibus, school bus, trolleybus \\
storefront & grocery store, tobacco shop, bookstore, toy store, barbershop, candy store, shoe store \\
bicycle & tricycle, mountain bike, tandem bicycle, unicycle
\end{longtable} 
\end{center}

\subsection{The Progress Gap between Standard and Real World Generalization}
\label{app:progress-gap}
\begin{figure}[t!]
\centering
\includegraphics[width=0.8\textwidth]{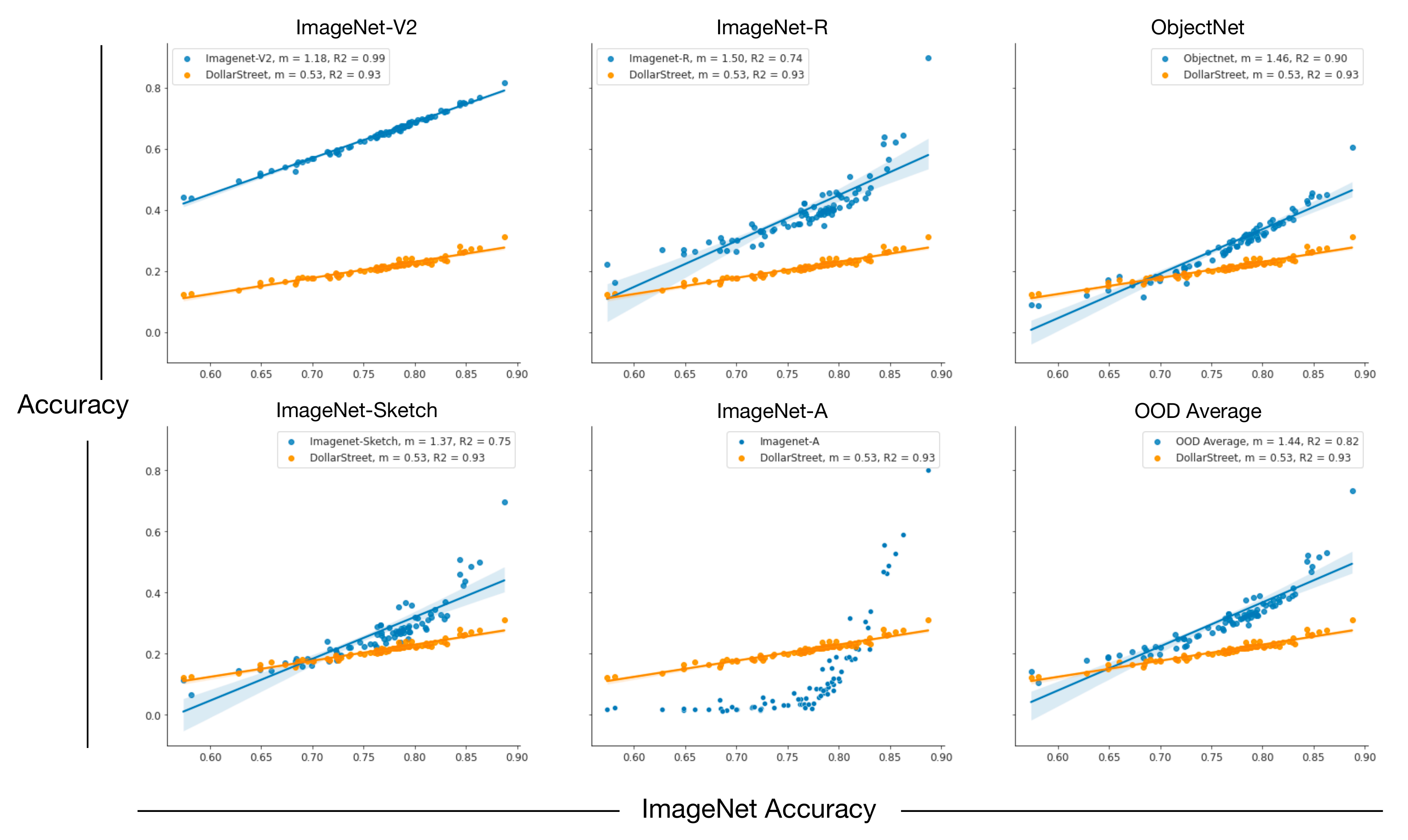}
\caption{Progress on each benchmark (blue) as a function of ImageNet, compared to DollarStreet (orange).} 
\label{fig:ds_progress_all}
\end{figure}

\begin{figure}[t!]
\centering
\includegraphics[width=0.8\textwidth]{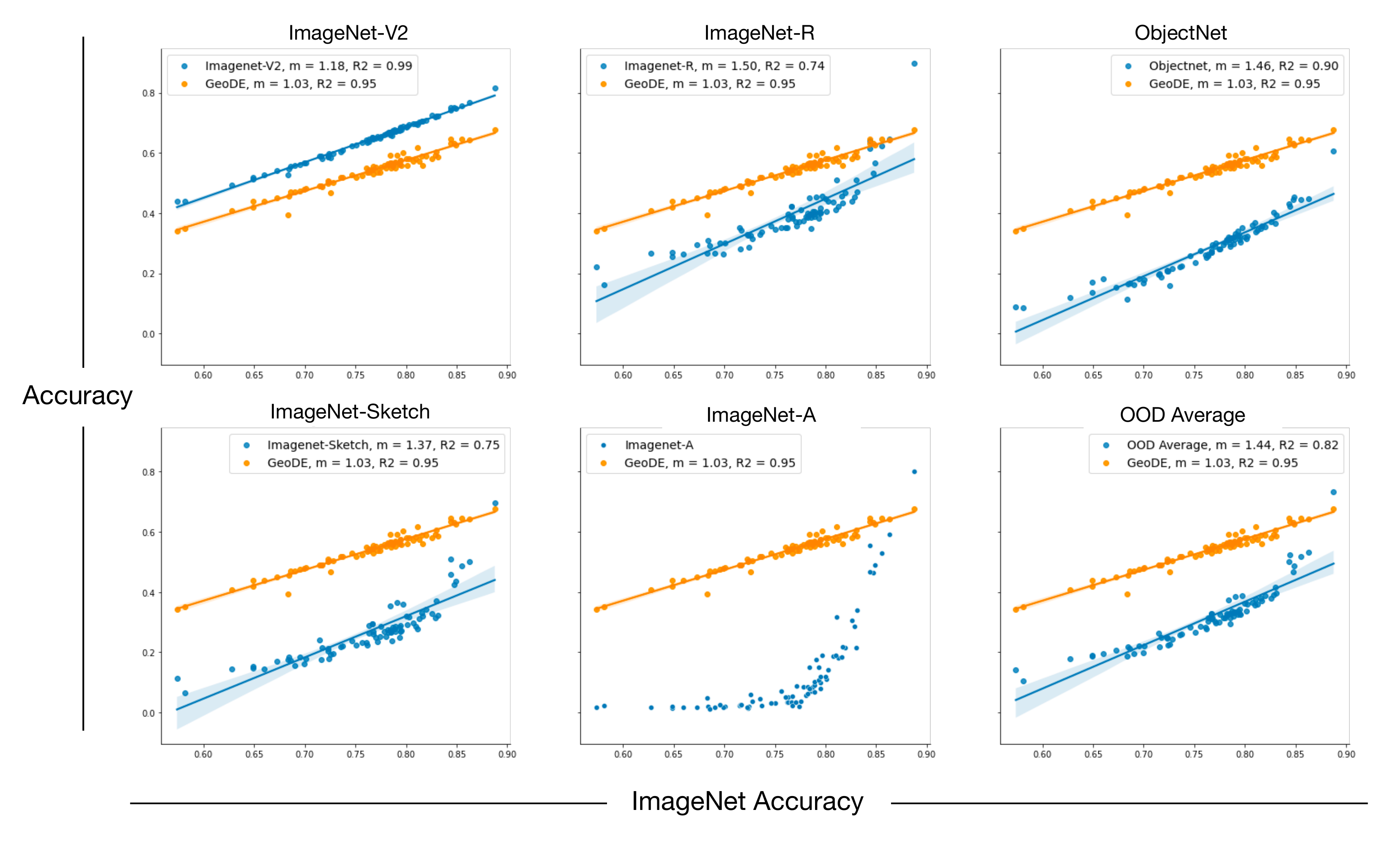}
\caption{Progress on each benchmark (blue) as a function of ImageNet accuracy, compared to GeoDE (orange).} 
\label{fig:geode_progress_all}
\end{figure}

In \Cref{fig:ds_progress_all} and \Cref{fig:geode_progress_all} we show the performance on each standard ImageNet benchmark as a function on ImageNet performance, comparing the progress rates with DollarStreet and GeoDE respectively. 

\subsection{Performance Disparities}
\label{app:performance-disparities}

\begin{figure}[t!]
\centering
\includegraphics[width=0.8\textwidth]{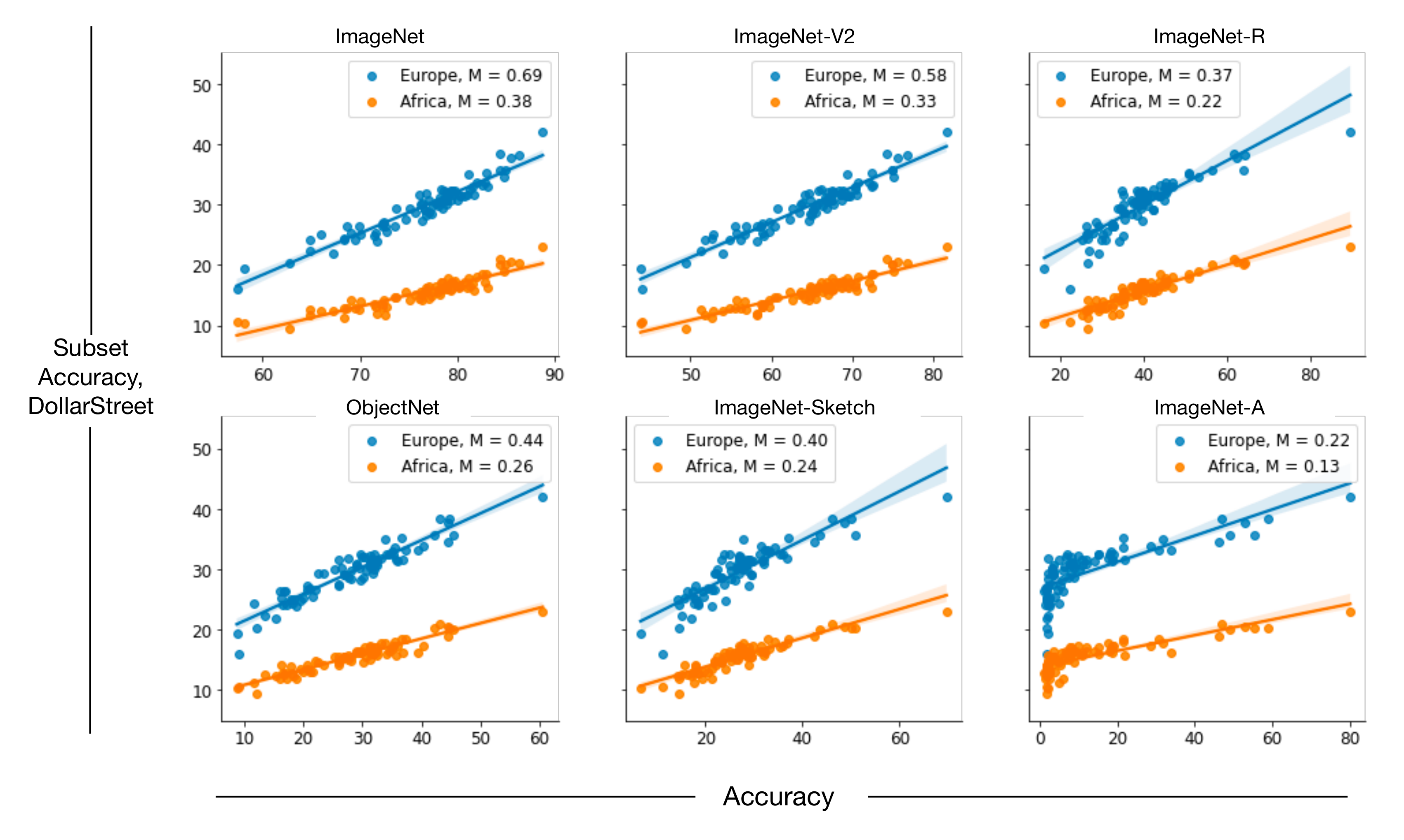}
\caption{\textbf{Model improvement on both in-distribution and out-of-distribution benchmarks exacerbates the region disparity on DollarStreet}. Region disparity is measured as the accuracy difference between Europe and Africa subsets.} 
\label{fig: ds_subset_all}
\end{figure}

\begin{figure}[t!]
\centering
\includegraphics[width=0.8\textwidth]{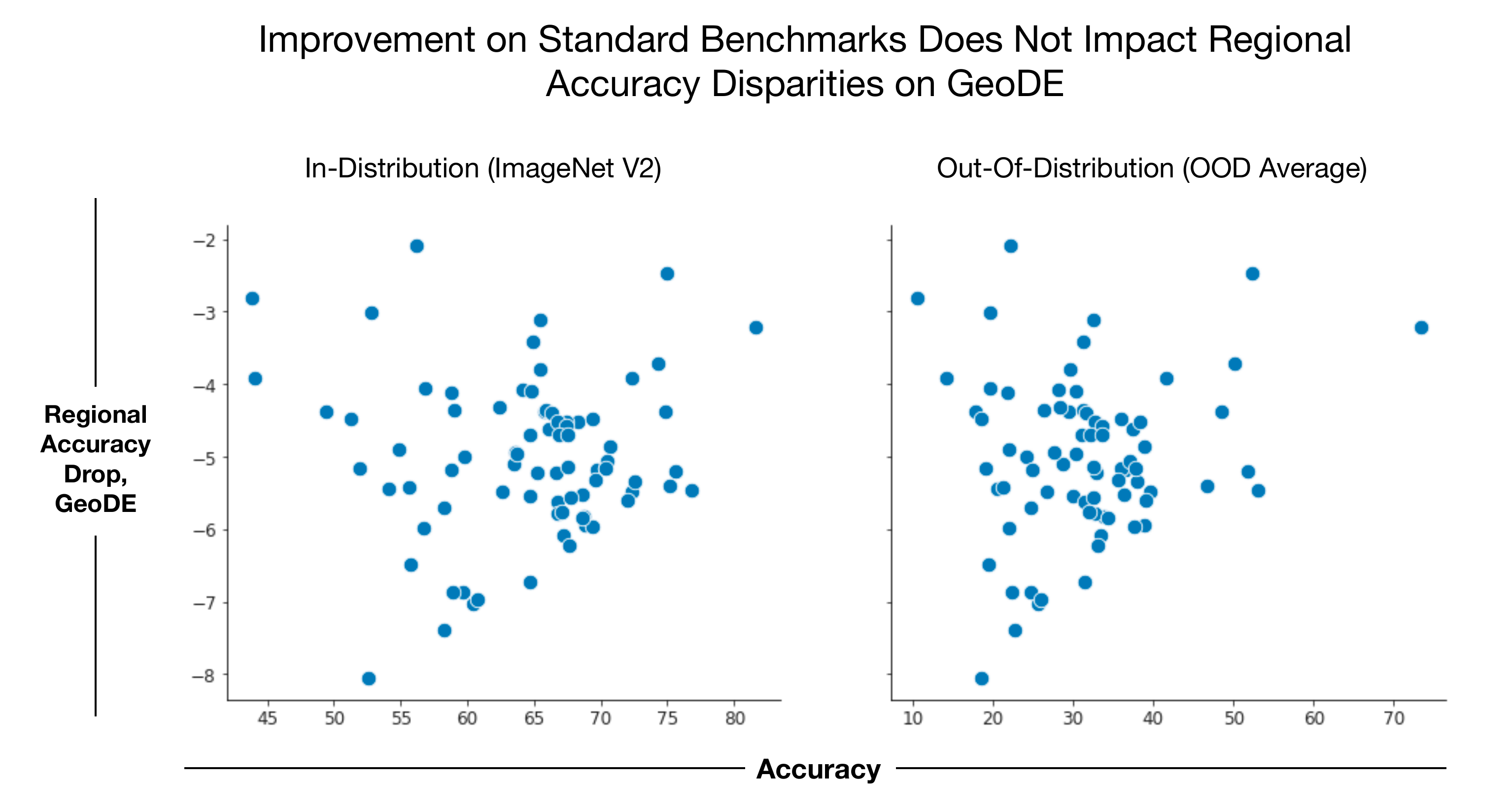}
\caption{\textbf{Model improvement on both in-distribution and out-of-distribution benchmarks fails to improve the region disparity on GeoDE}. Region disparity is measured as the accuracy difference between Europe and Africa subsets.} 
\label{fig:geode_grid}
\end{figure}

\begin{figure}[t!]
\centering
\includegraphics[width=0.8\textwidth]{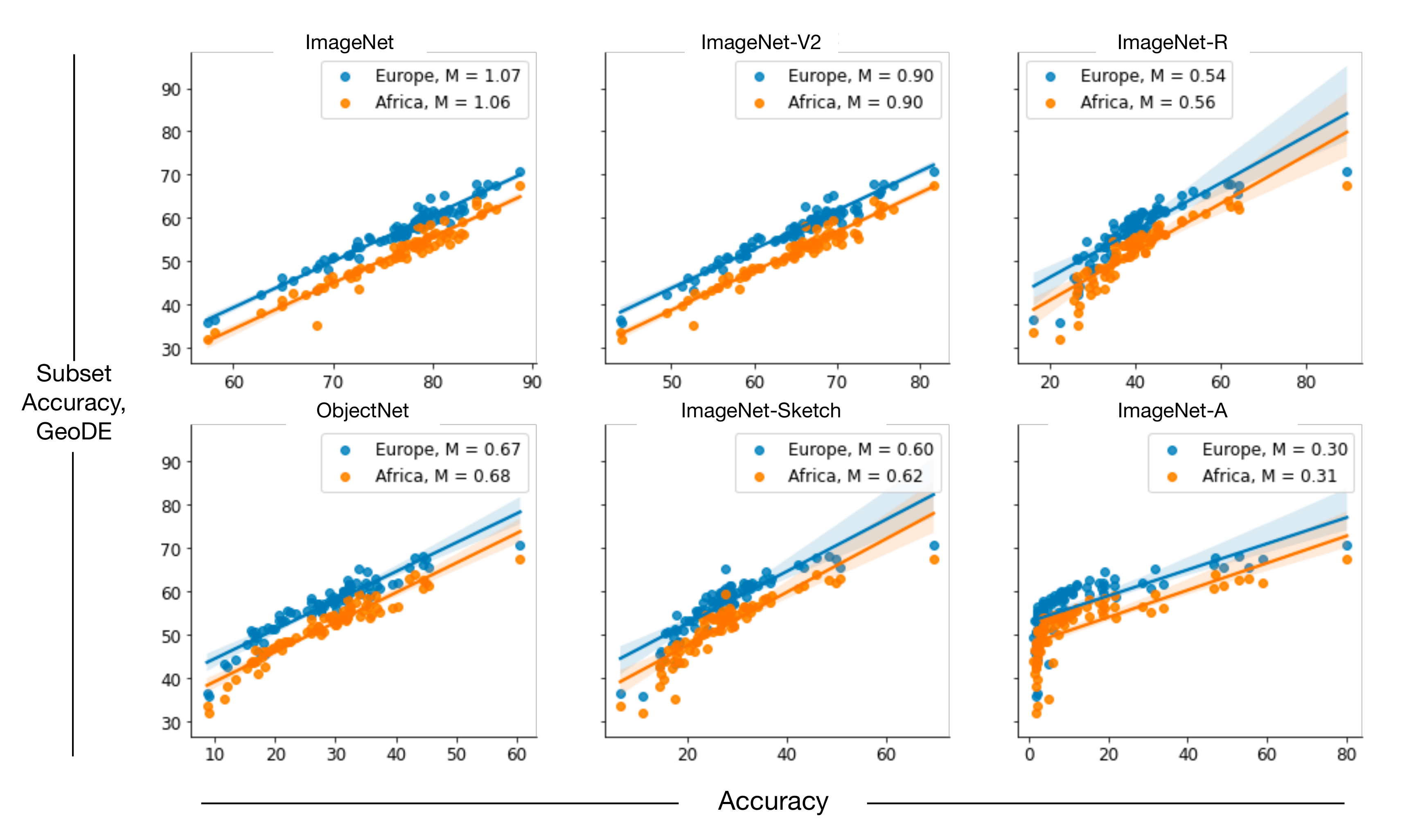}
\caption{Model improvement on both in-distribution and out-of-distribution benchmarks does not improve the region disparities on GeoDE.} 
\label{fig: geode_subset_all}
\end{figure}

We show the GeoDE version of \Cref{fig:dollarstreet_grid} below in \Cref{fig:geode_grid}, finding that improvement on standard imagenet benchmarks does not significantly impact regional accuracy disparities on GeoDE. We also show the relationships between Europe and Africa subsets of DollarStreet and GeoDE individually in \Cref{fig: ds_subset_all} and \Cref{fig: geode_subset_all}. 

\subsection{Foundation Models and Scaling}
\label{app:foundation_and_scaling}
\begin{figure}[htb]
\centering
\includegraphics[width=0.8\textwidth]{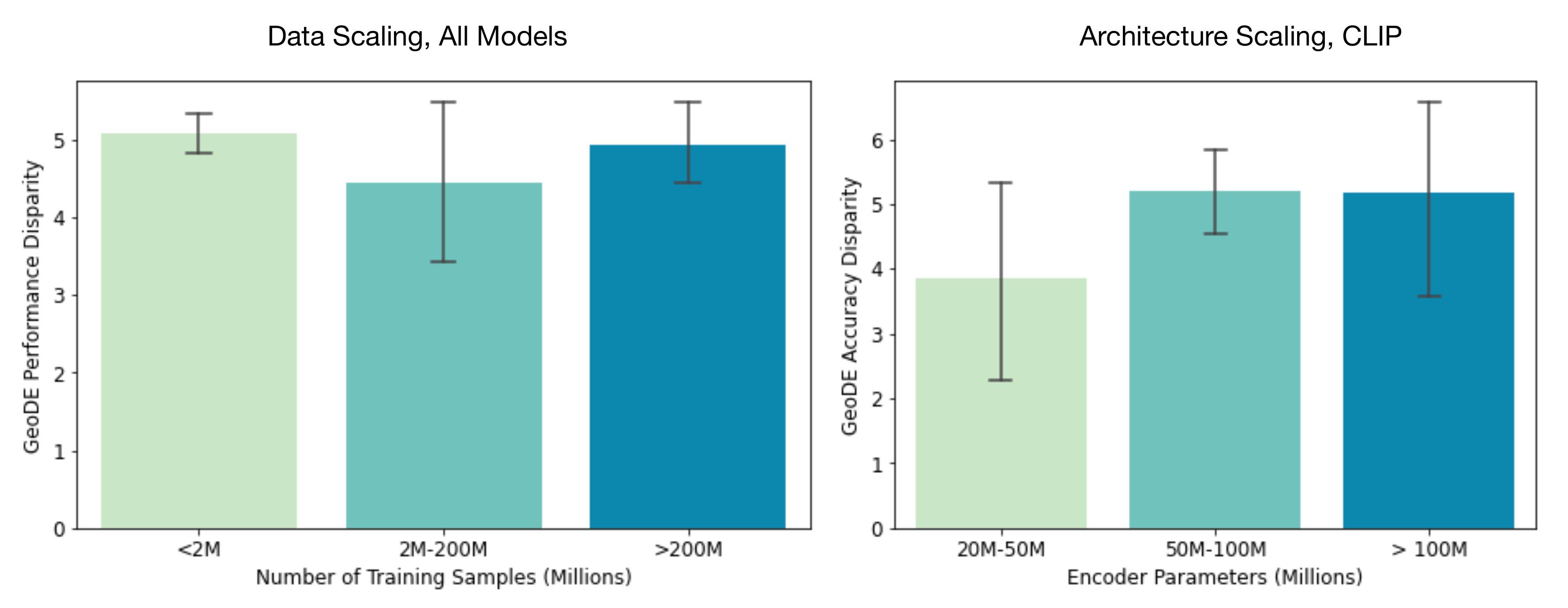}
\caption{\textbf{Dataset and architecture scaling fails to reduce region disparities on GeoDE.}}
\label{fig:geode_scaling}
\end{figure}

We replicate the plots in \Cref{fig:scale_clip} for GeoDE in \Cref{fig:geode_scaling}. 

\subsection{Representative Data}
\label{app:representative_data}

The GeoDE classes with overlapping ImageNet labels of DollarStreet include: hand soap, dustbin, chair, light fixture, light switch, plate of food, stove, cooking pot, cleaning equipment, lighter, medicine, toy, jug, house , waste container, car, road sign, wheelbarrow, storefront, bicycle.

%%%%%%%%%%%%%%%%%%%%%%%%%%%%%%%%%%%%%%%%%%%%%%%%%%%%%%%%%%%%
\end{document}